\documentclass[10pt,twocolumn,letterpaper]{article}

\usepackage[pagenumbers]{cvpr} 

\usepackage[dvipsnames]{xcolor}

\newcommand{\ours}[0]{DeDoDe}
\newcommand{\A}[0]{\ensuremath{\mathcal{A}}}
\newcommand{\B}[0]{\ensuremath{\mathcal{B}}}

\usepackage[style=english]{csquotes}
\usepackage[detect-all,per-mode=symbol,group-digits=integer,group-minimum-digits=4,group-separator={,}]{siunitx}
\newcommand{\parsection}[1]{\vspace{0mm}\noindent\textbf{#1}~}

\usepackage[pagebackref,breaklinks,colorlinks]{hyperref}

\def\paperID{7} 
\def\confName{3DV}
\def\confYear{2024}

\title{DeDoDe: Detect, Don't Describe --- Describe, Don't Detect \\for Local Feature Matching}

\author{Johan Edstedt$^1$
\quad
Georg Bökman$^2$
\quad
 Mårten Wadenbäck$^1$
\quad
 Michael Felsberg$^1$
 \\
{\normalsize $^1$Linköping University,
$^2$Chalmers University of Technology}
}
\begin{document}
\maketitle
\begin{abstract}
Keypoint detection is a pivotal step in 3D reconstruction, whereby sets of (up to) K points are detected in each view of a scene. Crucially, the detected points need to be consistent between views, i.e., correspond to the same 3D point in the scene.
One of the main challenges with keypoint detection is the formulation of the learning objective.
Previous learning-based methods typically jointly learn descriptors with keypoints, and treat the keypoint detection as a binary classification task on mutual nearest neighbours. 
However, basing keypoint detection on descriptor nearest neighbours is a proxy task, which is not guaranteed to produce 3D-consistent keypoints. Furthermore, this ties the keypoints to a specific descriptor, complicating downstream usage.
In this work, we instead learn keypoints directly from 3D consistency. 
To this end, we train the detector to detect tracks from large-scale SfM. As these points are often overly sparse, we derive a semi-supervised two-view detection objective to expand this set to a desired number of detections. To train a descriptor, we maximize the mutual nearest neighbour objective over the keypoints with a separate network.
Results show that our approach, \ours, achieves significant gains on multiple geometry benchmarks. Code is provided at \href{https://github.com/Parskatt/DeDoDe}{github.com/Parskatt/DeDoDe}. \end{abstract}
    
\section{Introduction}
\label{sec:intro}

Incremental Structure from Motion (SfM) can roughly be divided into detection, matching, and estimation. In the detection stage, we seek to detect keypoints in the 3D scene that are consistent between views. This is a crucial step, as the optimization process relies on accurate 3D tracks, \ie, 3D points consistently detected in multiple views.
While keypoints are necessary for SfM, it is difficult to formulate a learning-based objective. Traditional methods often rely on proxy tasks such as extrema in differences of Gaussians~\cite{lowe2004distinctive}, corner detection~\cite{harris1988combined}, while learning-based methods often aim to detect mutual nearest neighbours of descriptors~\cite{tyszkiewicz2020disk, gleize2023silk}. While these methods can result in distinct and reproducible points, they do not directly optimize 3D consistency of the points between views.
\begin{figure}
    \centering
    \includegraphics[width = .95\linewidth]{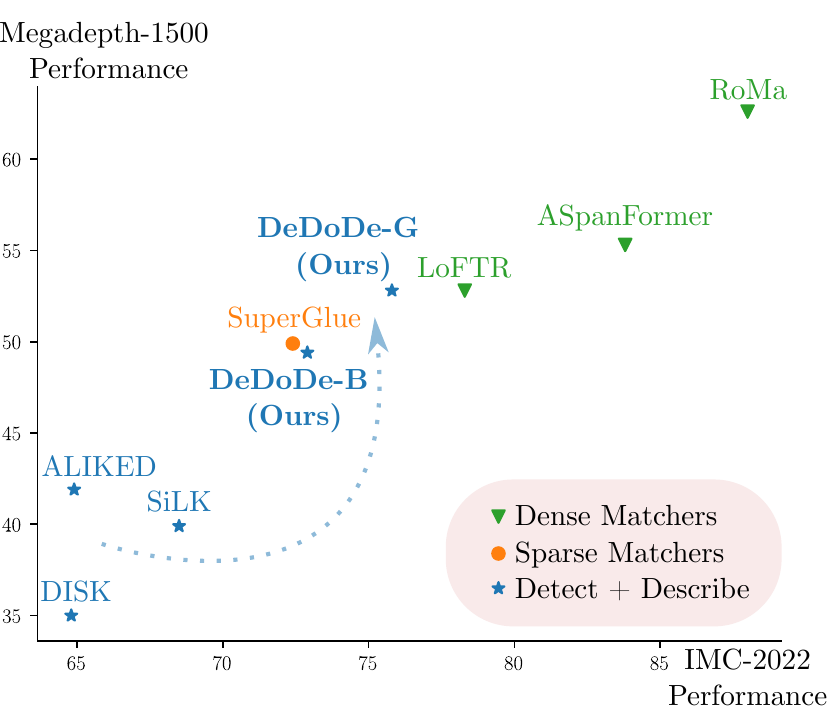}
    \caption{\textbf{\ours~closes the matching gap.} There is a large gap in performance between traditional detector/descriptor models and fully-fledged matching models. \ours~significantly reduces this gap, demonstrating the untapped potential of the simple detector+descriptor approach.}
    \label{fig:teaser-figure}
\end{figure}
In order to tackle this discrepancy, we propose to directly learn 3D consistency by detecting 3D tracks. To do this we use SfM reconstructions directly as supervision.
\begin{figure*}
    \centering
    \includegraphics[height = .18\linewidth]{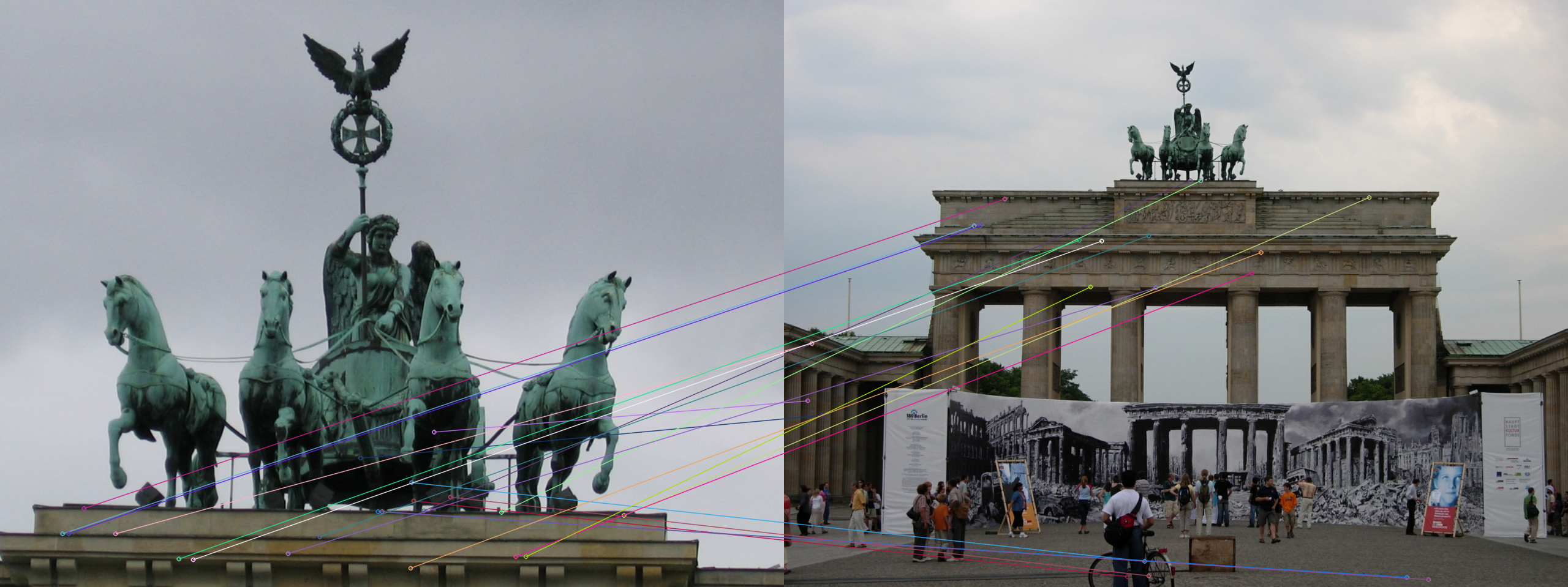}
    \includegraphics[height = .18\linewidth]{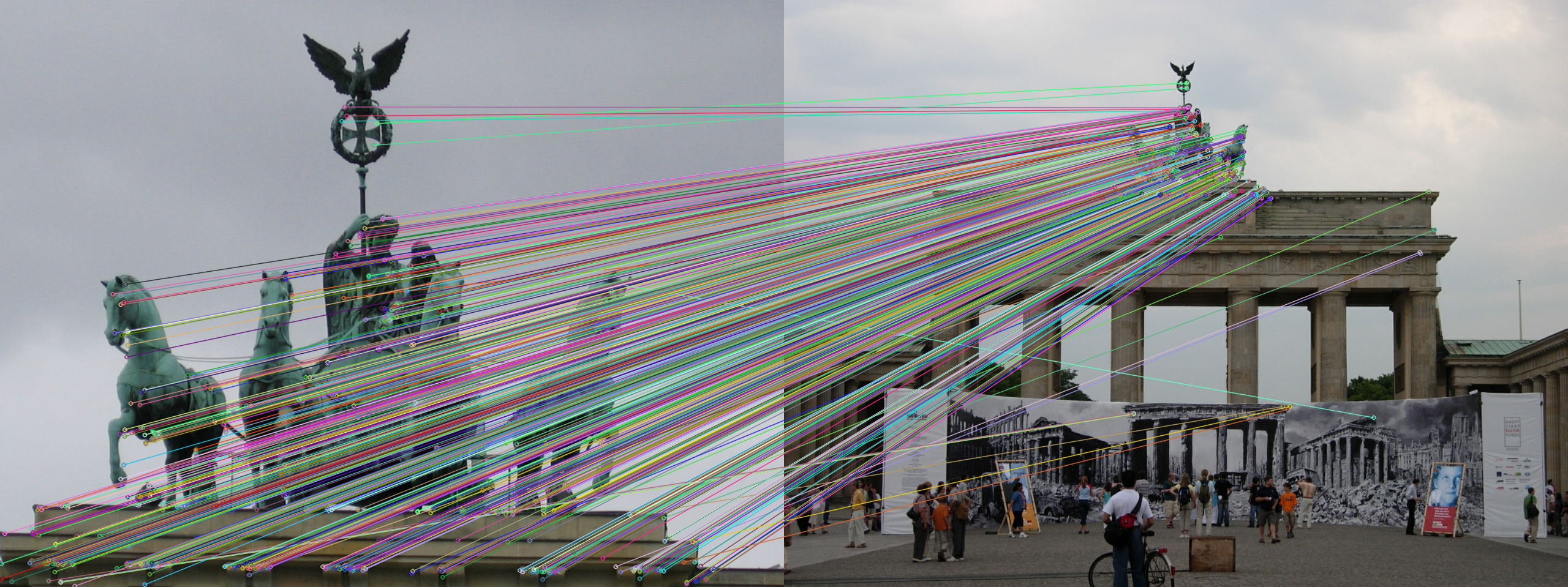}
    \caption{\textbf{Qualitative comparison.} \textbf{Left:} Matches from the joint detector/descriptor DISK~\cite{tyszkiewicz2020disk}, \textbf{Right:} Matches from our decoupled detector/descriptor \ours. \ours~shows substantial improvements in both detector precision and descriptor quality in comparison to previous approaches, performing well even under significant changes in viewpoint.}
    \label{fig:qualitative-comparison}
\end{figure*}
Since we have access to ground truth 3D tracks, we can directly optimize to detect these. However, as the 3D tracks are produced by a base detector, there is a risk of missing reliable keypoints discarded by the original detector. 
To mitigate this problem, we propose a semi-supervised two-view top-$k$ consistency objective, which enables us to surpass the initial detector both in precision and recall. A detailed description of our approach is provided in Section~\ref{sec:detector}. 

When training detectors and descriptors jointly, the descriptor objective is complex. In contrast, we can directly optimize the mutual nearest neighbour negative log-likelihood for the descriptions over \ours~keypoints to train the \ours~descriptor, which we will refer to as \ours-B. While this model performs well, local features struggle with describing repeating structures, which requires a larger context. Recent work~\citep{edstedt2023roma} has shown promising results of using globally informed DINOv2~\cite{oquab2023dinov2} features for dense matching. We show that incorporating frozen DINOv2 features into a larger version of our descriptor, which we call \ours-G, also increases performance. Our approach to descriptor learning is detailed in Section~\ref{sec:descriptor}. 

Finally, we perform an extensive set of SotA experiments and ablations in Section~\ref{sec:experiments}, in which we find that \ours~sets a new state-of-the-art in feature matching. Our approach bridges the gap between the traditional detector+descriptor approach to matching and modern end-to-end matchers as can be seen in Figure~\ref{fig:teaser-figure}. A qualitative comparison between the joint detector/descriptor DISK~\cite{tyszkiewicz2020disk} and \ours~is presented in Figure~\ref{fig:qualitative-comparison}.
\begin{figure*}
    \centering
    \includegraphics[width = .99\linewidth]{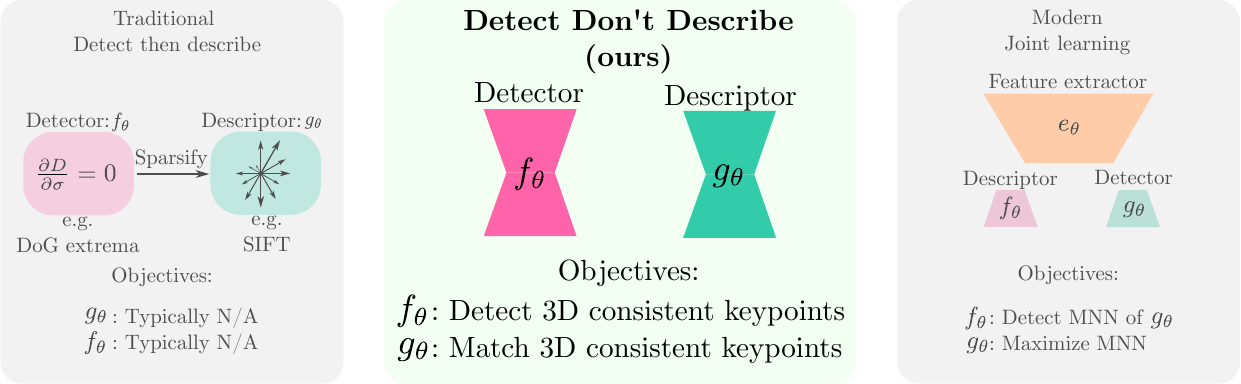}
    \caption{\textbf{Previous detector/descriptor approaches compared to \ours.} \textbf{Left:} Traditional methods are typically two-stage, first detecting promising keypoints with a detector $f_{\theta}$, which are subsequently described by a descriptor $g_{\theta}$. This approach suffers from the lack of synergy between the detector and descriptor~\cite{taira2018inloc,dusmanu2019d2}. \textbf{Right:} Later, learning-based methods using the joint learning paradigm were introduced. Here the objectives of the descriptor and the detector are aligned. However, this has other issues. One issue is that introducing inter-dependencies of the descriptor/detector often degrades performance~\cite{li2022decoupling}. For example, repeating complex patterns may constitute excellent keypoints, but fail to match, while unique textureless regions may match, but do not produce exact localization. \textbf{Middle:} In this paper we take a new approach and decouple the descriptor and detector learning while having aligned objectives. We do this by learning to detect and match 3D tracks from large-scale SfM. In Section~\ref{sec:detector} we describe how we create a training objective for the detector, and in Section~\ref{sec:descriptor} we describe the approach for the descriptor. Our approach leads to significant improvements to the state-of-the-art, as demonstrated in Section~\ref{sec:experiments}, showing the potential of detector based methods.}
    \label{fig:method}
\end{figure*}
\section{Related Work}

\parsection{Keypoint Detection from Descriptors:}
\citet{dusmanu2019d2} proposed a joint descriptor/detector pipeline. \citet{revaud2019r2d2} extended these ideas and proposed a set of auxiliary losses to improve the detector. \citet{tyszkiewicz2020disk} proposed DISK, which uses a joint training objective for keypoint detection and description through policy gradient optimization, enabling backpropagation to the detector. \citet{zhao2022alike,Zhao2023ALIKED} proposed ALIKE and ALIKED, where a local softmax weighting approach is used for keypoint detection that enables backpropagation to the detector.

\parsection{Keypoint Detection from a Base Detector:}
Closest to our work are keypoint detectors that aim to surpass a base detector. One of the earliest works TaSK~\cite{strecha2009training} classified keypoints that are consistent over long time periods, an idea that was further improved in TILDE~\cite{verdie2015tilde}. A similar approach was investigated by \citet{hartmann2014predicting}, where a classifier was trained to discriminate SfM matchable keypoints. LIFT~\cite{yi2016lift} crops small patches around SfM tracks, but does not optimize for detecting tracks directly.
SuperPoint~\cite{detone2018superpoint} is trained from an initial base detector by means of a synthetic training set of corners. A second detector is then trained to produce consistent detections of the base detector for heavily augmented images. More recent approaches, such as SOLD$^2$~\cite{pautrat2021sold2} and DeepLSD~\cite{pautrat2023deeplsd} further develop these ideas for line detection. Our work can be seen as further development of these ideas with some key differences. In contrast to previous work we introduce a top-$k$ two-view semi-supervised detection objective combined with a coverage objective. Furthermore, we use an aligned but decoupled detector/descriptor learning approach. See Section~\ref{sec:detector}. 

\parsection{Whether to Couple or Decouple Keypoint Detection and Description:}
Traditionally, handcrafted feature detection pipelines, \eg SIFT~\cite{lowe2004distinctive}, the keypoint detection is independent of the descriptors.
Early learned work in a similar fashion followed the decoupled approach~\cite{mishchuk2017working,mishkin2018repeatability,barroso2019key}.
However, this approach has been called into question in later works~\cite{sattler2018benchmarking, taira2018inloc, dusmanu2019d2}, with the reasoning that descriptors often succeed in matching where keypoints fail.
Hence most following work has worked in the joint learning setting~\cite{revaud2019r2d2, tyszkiewicz2020disk, gleize2023silk}, which binds the keypoints to a specific descriptor. Recently, \citet{li2022decoupling}, argue that the joint training is detrimental, as poor descriptors lead to unreliable mutual nearest neighbours, resulting in worse keypoint detectors.
However, they still use the descriptor features as input to the detector network.
In contrast to joint previous works, we consider an aligned objective that is still decoupled.
\section{Why Decouple Keypoint Detection and Description?}
In this paper we argue for a descriptor-agnostic approach to keypoint detection, see Figure~\ref{fig:method}. There are multiple reasons for this, we summarize the main points below
\begin{enumerate}
    \item \textbf{Compatability:} Mutual nearest neighbours of a certain descriptor does not guarantee general matchability or consistency. By being agnostic to the descriptor, our keypoints can be used for arbitrary matchers, as we demonstrate in Table~\ref{tab:megadepth-loftr}.
    \item \textbf{Decoupling Gains:} Thinking in a modular fashion lets us design decoupled experiments, for example Table~\ref{tab:detect-repeatability-10000} and~\ref{tab:detect-repeatability-2000}, as well as Table~\ref{tab:ablation}.
    \item \textbf{Performance:} Our approach improves significantly on the state-of-the-art, see Figure~\ref{fig:teaser-figure}. This demonstrates the soundness of our approach.
\end{enumerate}
The following sections describe our method. In Section~\ref{sec:detector} we introduce the \ours~detector, and in Section~\ref{sec:descriptor} the \ours~descriptor.

\section{Detect, Don't Describe}
\label{sec:detector}
\begin{figure*}[t!]
    \centering
    \includegraphics[width = .99\linewidth]{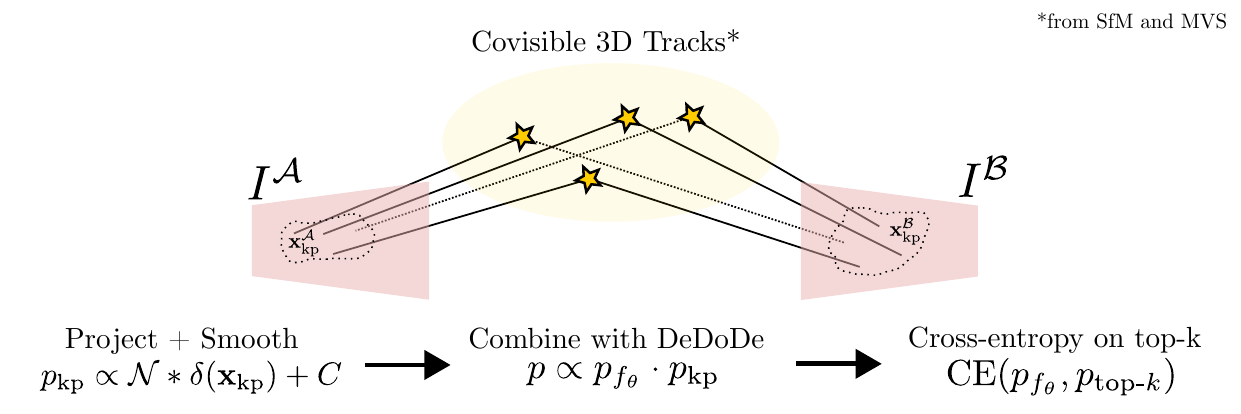}
    \caption{\textbf{Training objective of \ours~for keypoint detection.} We sample an image pair from a large-scale SfM + MVS reconstruction. (i) For this pair we find all covisible detections (from a base detector) that resulted in a 3D track. We take the union of tracks detected in the image (indicated by filled lines in the figure), and tracks not detected but covisible (indicated by dotted lines in the figure). See Section~\ref{sec:kpts-as-tracks}. (ii) We project these detections into the images. We then convert this point set into a softened distribution, which constitutes our prior for good keypoints, see Section~\ref{sec:prior-detection-dist}. (iii) We then condition this distribution with the predictions of \ours, as the base detector may have false negatives, see Section~\ref{sec:posterior-detection-dist}. (iv) Finally, we threshold this distribution at the $k$:th highest value, and compute the cross-entropy between \ours~and the target. Using top-k stabilizes the self-supervised objective. See Section~\ref{sec:detector-training-objective}.}
    \label{fig:detector}
\end{figure*}

For an overview of our approach, see Figure~\ref{fig:detector}.

\subsection{Preliminaries}
Let $f_{\theta}(x|I)$ be a mapping producing a log density (possibly non-normalized) over an image $I$. 
We would like this network to produce high values for all 3D consistent keypoints, and low values otherwise. Now let 
\begin{equation}
\mathcal{D} =\{I^j\}_{j=1}^{J}    
\end{equation}
 be a dataset of images with cardinality $|\mathcal{D}| = J$. We will denote the set of keypoints for a given image $I^j$ in pixel coordinates as $\textbf{x}^j_{\rm kp} = \{x^j_{i}\}_{i=1}^{K^j}$ where $K^j$ is the number of observable keypoints in image $I^j$. Formally, we want to maximize the following log-likelihood objective,

\begin{equation}
\label{eq:objective}
    \max_{\theta} \sum_{j = 1}^{|\mathcal{D}|} \sum_{i=1}^{K^j} f_{\theta}(x_{i}^j|I^j) - \log Z_{\theta}(I^j),
\end{equation}
where we have defined
\begin{equation}
    Z_{\theta}(I^j) := \sum_{x^{j} \in I^j} \exp (f_{\theta}(x^j|I^j)).\footnote{The notation $x^{j} \in I^j$ here means normalization over \emph{all} pixel coordinates in image $I^j$.}
\end{equation}

That is, we want to maximize the likelihood of sampling 3D keypoints. For notational convenience, we will also define
\begin{equation}
    p_{f_{\theta}}(x_{i}^{j}) := \exp (f_{\theta}(x_{i}^{j}|I^j) - \log Z_{\theta}(I^j))
\end{equation}

The main challenge of keypoint detection is how to maximize Equation~\ref{eq:objective}. In particular, in contrast to other keypoint tasks, there are no ground truth annotations. In this paper we will consider a good prior for a keypoint as a 2D detection (from a base keypoint detector) that has \enquote{survived} 3D reconstruction, \ie, formed a 3D track.

\subsection{Keypoints as 3D Tracks}
\label{sec:kpts-as-tracks}
We use a dataset of large-scale SfM reconstructions, MegaDepth~\cite{li2018megadepth}, where SIFT~\cite{lowe2004distinctive} is used as keypoint detector, \ie, scale space extrema. These detections go through a filtering process, whereby only 3D consistent detections remains after optimization finishes. However, in each image, only a fraction of all visible tracks are detected. To solve this issue we propose to sample pairs of images $I^{\A}$ and $I^{\B}$, and use the union of covisible detections. We obtain the covisibility by means of Multi-View Stereo depth maps. 

\subsection{A Smooth Keypoint Detection Prior}
\label{sec:prior-detection-dist}
Due to the detector producing detections on a discrete grid, and for reasons we will discuss in Section~\ref{sec:posterior-detection-dist}, a smooth detection prior is desirable. We do this by a 2-stage process. For both $I^{\A}$ and $I^{\B}$ we start by placing Dirac deltas $\delta(\mathbf{x}_{\rm kp})$ at the nearest grid position of the detections. We then convolve these with a Gaussian kernel with $\sigma = 0.5$ pixels, and add a small constant that corresponds to a uniform distribution.  Mathematically,
\begin{align}
\label{eq:prior1}
    \Tilde{p}_{\rm kp}^{\A} \propto \mathcal{N}(0,\sigma^2) * \delta(\mathbf{x}_{\rm kp}^{\A}) + C^{\A},\\
    \label{eq:prior2}
        \Tilde{p}_{\rm kp}^{\B} \propto \mathcal{N}(0,\sigma^2) * \delta(\mathbf{x}_{\rm kp}^{\B}) + C^{\B}.
\end{align}
Finally we warp the distributions of $\A, \B$ using the depth to produce $\Tilde{p}_{\rm kp}^{\A \to \B}, \Tilde{p}_{\rm kp}^{\B \to \A}$. Combining yields  
\begin{align}
    p^{\A}_{\rm kp} \propto \Tilde{p}_{\rm kp}^{\A} \cdot \Tilde{p}_{\rm kp}^{\B \to \A},\\
    p^{\B}_{\rm kp} \propto \Tilde{p}_{\rm kp}^{\B} \cdot \Tilde{p}_{\rm kp}^{\A \to \B}.
    \end{align}

In practice, the constants in equations~\ref{eq:prior1} and \ref{eq:prior2} are implicitly defined by approximating $\log p_{\rm kp} \approx \vartheta p_{\rm kp}$, with $\vartheta=50$.
The end result is a detection prior which produces particularly high values at 3D-tracks detected in \emph{both} images, which we found beneficial. 

Note that while this is a simple and powerful way of producing a detection prior, there is a multitude of ways one could construct it.
For example, we could exclude all tracks smaller than a certain track length, or only include tracks with sufficiently small reprojection errors.
However, such approaches introduce more complex hyperparameters, and we found our simple approach to already yield state-of-the-art results.

\subsection{Posterior Detection Distribution}
\label{sec:posterior-detection-dist}
While the detection prior yields good detections, we found that it is often overly sparse in practice. This is due to the base detector not detecting all keypoints, \ie, insufficient recall. To amend this issue, we propose to update the prior distribution with the predictions of $f_{\theta}$, \ie,
\begin{equation}
    p \propto p_{f_{\theta}} \cdot p_{\rm kp}.
\end{equation}
This enables the network to detect keypoints missed by the base detector. Note that while it would appear that multiplying the distributions cannot add detections, the fact that we have put a small flat prior on the keypoint distribution in practice creates peaks around the model detections that enable detection of additional keypoints.
\subsection{Detector Training Objective}
\label{sec:detector-training-objective}
In principle, one could form an objective directly using the cross entropy between $p_{f_{\theta}}$ and $p$. However, we found training to be more stable when thresholding and binarizing $p$ in practice. This is due to the non-binary objective having degenerate solutions, as we show in Suppl.~\ref{sec:why-topk}. To this end, we select the top-$k$ detections in a batch, where we empirically select $k = {\rm batch size} \cdot 1024$, to construct the target distributions. Then we simply compute the cross entropy between the predicted distribution and the target as a loss,
\begin{equation}
    \mathcal{L}_{\rm detection} = {\rm CE}(p_{f_{\theta}}, p_{\text{top-}k}).
\end{equation}

We additionally add a coverage regularization on $f_{\theta}$ by computing the cross-entropy between a low-pass filtered $p_{f_{\theta}}$ and a binary distribution $p_{\rm MVS}$ indicating successful MVS reconstruction, \ie,
\begin{equation}
    \mathcal{L}_{\rm coverage} = {\rm CE}(\mathcal{N}(0,\sigma^2) * p_{f_{\theta}}, \mathcal{N}(0,\sigma^2) * p_{\rm MVS}).
\end{equation}
In practice we used $\sigma = 12.5$ pixels.
This regularization is necessary, as the network may otherwise produce detections in unmatchable regions (\eg sky).
Our combined loss is
\begin{equation}
    \mathcal{L} = \mathcal{L}_{\rm detection} + \mathcal{L}_{\rm coverage}.
\end{equation}

As with our choice of prior distribution, these choices are not guaranteed to be optimal, but worked well in practice. Note that the choice of $k$ depends on the expected overlap between the views. For difficult pairs, one would expect fewer keypoints to overlap. It is in principle possible to control the sparsity of the detector by decreasing or increasing $k$, but investigation of this is outside the scope of the paper.

\subsection{Sampling Keypoints at Inference}
To sample keypoints, we follow previous work and select the top $K$ most likely keypoints from $f_{\theta}$. We follow SiLK~\cite{gleize2023silk} and do not employ non-max supression.
\subsection{Network Architecture}
\parsection{Encoder:}
We use a VGG-19 network pretrained on ImageNet, implemented in \texttt{torchvision}. We use strides $[1,2,4,8]$. The output at each scale is a feature-map with $[64,128,256,512]$ channels respectively.

\parsection{Decoder:}
We use the architecture of the depthwise convolution refiners proposed in DKM~\cite{edstedt2023dkm}, with 8 blocks per scale and internal dimension $[64,128,256,512]$.
The input at each scale is the encoded features, as well as context from previous decoding layers. The output at each scale is a residual addition to a dense grid of keypoint logits. Between each scale these logits are upsampled using bicubic interpolation.

Note that our network is different from DKM which takes two images as input, constructing a local correlation volume + stacked feature maps, while ours simply refines the features.

\section{Describe, Don't Detect}
\label{sec:descriptor}
\subsection{Descriptor Preliminaries}
Given two sets of keypoints, $\textbf{x}^{\A}_{\rm kp}$, $\textbf{x}^{\B}_{\rm kp}$, and the respective images $I^{\A}, I^{\B}$, the task of a descriptor network $\mathbf{g}_{\theta}(x|I^{\A})$ is to produce descriptions $\mathbf{g}_{\theta}(\textbf{x}^{\A}_{\rm kp}|I^{\A}), \mathbf{g}_{\theta}(\textbf{x}^{\B}_{\rm kp}|I^{\B})$ that match well for keypoints corresponding to the same 3D point, and match poorly for non-corresponding keypoints.

We choose the probabilistic framework, as is common~\cite{tyszkiewicz2020disk, gleize2023silk}, where
\begin{equation}
    p_{g_{\theta}}(x^{\A} | x^{\B}) := \frac{\exp (\langle \mathbf{g}_{\theta}(x^{\A}|I^{\A}), \mathbf{g}_{\theta}(x^{\B}|I^{\B})\rangle )}
    {\sum_{x_i^{\A} \in \textbf{x}^{\A}_{\rm kp}}\exp (\langle \mathbf{g}_{\theta}(x_i^{\A}|I^{\A}), \mathbf{g}_{\theta}(x^{\B}|I^{\B})\rangle)},
\end{equation}
defines the conditional matching distribution,
and formally define matching likelihoods as
\begin{equation}
\label{eq:descriptor}
    L_{g_{\theta}}(x^{\A}, x^{\B}) := p_{g_{\theta}}(x^{\B} | x^{\B}) \cdot p_{g_{\theta}}(x^{\B} | x^{\A}).
\end{equation}

This objective has the benefit of working well with mutual nearest neighbours based downstream matchers.

\subsection{Descriptor Training Objective}
Taking the logarithm of equation~\eqref{eq:descriptor} leads to a straightforward log-likelihood objective
\begin{equation}
\label{eq:descriptor-loss}
    \ell_{g_{\theta}}(x^{\A}, x^{\B}) = \log p_{g_{\theta}}(x^{\A} | x^{\B}) + \log p_{g_{\theta}}(x^{\B} | x^{\A}).
\end{equation}

The main difficulty with optimizing this objective is the normalizing term, which in most works is either approximated~\cite{tyszkiewicz2020disk}, or brute-force optimized in lower resolution~\cite{gleize2023silk}. However, as our formulation is decoupled, we face no such issues and simply use trained \ours~keypoints. During training, similarly to inference, we sample $K=5000$ descriptions from a dense grid, using a pretrained \ours~detector.

\subsection{Descriptor Network Architecture}
The architecture of the descriptor largely matches that of the detector. Importantly, however, they do not share any weights.

\parsection{Encoder:}
We use a VGG-19 network pretrained on ImageNet, implemented in \texttt{torchvision}. We use features at strides $[1,2,4,8]$. The output at each scale is a feature-map with $[64,128,256,512]$ channels respectively.

\parsection{Decoder:}
We use the architecture of the depthwise convolution refiners proposed in DKM~\cite{edstedt2023dkm}, with 5 blocks per scale and internal dimension $[32,64,256,512]$. We call this model \ours-B. The input at each stride is the encoded features, as well as context from previous decoding layers. The output at each scale is a residual addition to a dense grid of descriptions. Between each scale the descriptions are upsampled using bilinear interpolation. We use a description dimension of 256. We normalize the descriptions, and use a temperature of $20^{-1}$ during both training and inference.

Note, as for the detector, that our network is different from DKM which takes two images as input, constructing a local correlation volume + stacked feature maps, while ours simply refines the features of one image.

\parsection{\ours-G:}
To explore the full potential of detector descriptor methods, we propose to incorporate features from DINOv2~\cite{oquab2023dinov2}. This addition is simple in our framework, we just add frozen DINOv2 features at stride 14, with a corresponding decoder stage. We use an internal dimension of $768$ for this decoder. With this simple addition, we achieve impressive performance gains as demonstrated in Table~\ref{tab:megadepth-loftr} and Table \ref{tab:imc2022}.

\section{Experiments}
\label{sec:experiments}
\subsection{Training}

\parsection{\ours~Detector:}
We train for \num{100000} steps of batch size 8 on the MegaDepth dataset, using the same training split as DKM~\cite{edstedt2023dkm}. We use a fixed image size of $512\times 512$. Training is done on a single A100 GPU, and takes about 30 hours.

\parsection{\ours~Descriptor:}
Mirroring our detector training setup, we train for \num{100000} steps of batch size 8 on the MegaDepth dataset. We use a single A100 GPU, and training takes about 24 hours. The slightly shorter training time comes from using fewer blocks in the decoder.
\begin{figure*}
    \centering
    \includegraphics[width=.95\linewidth]{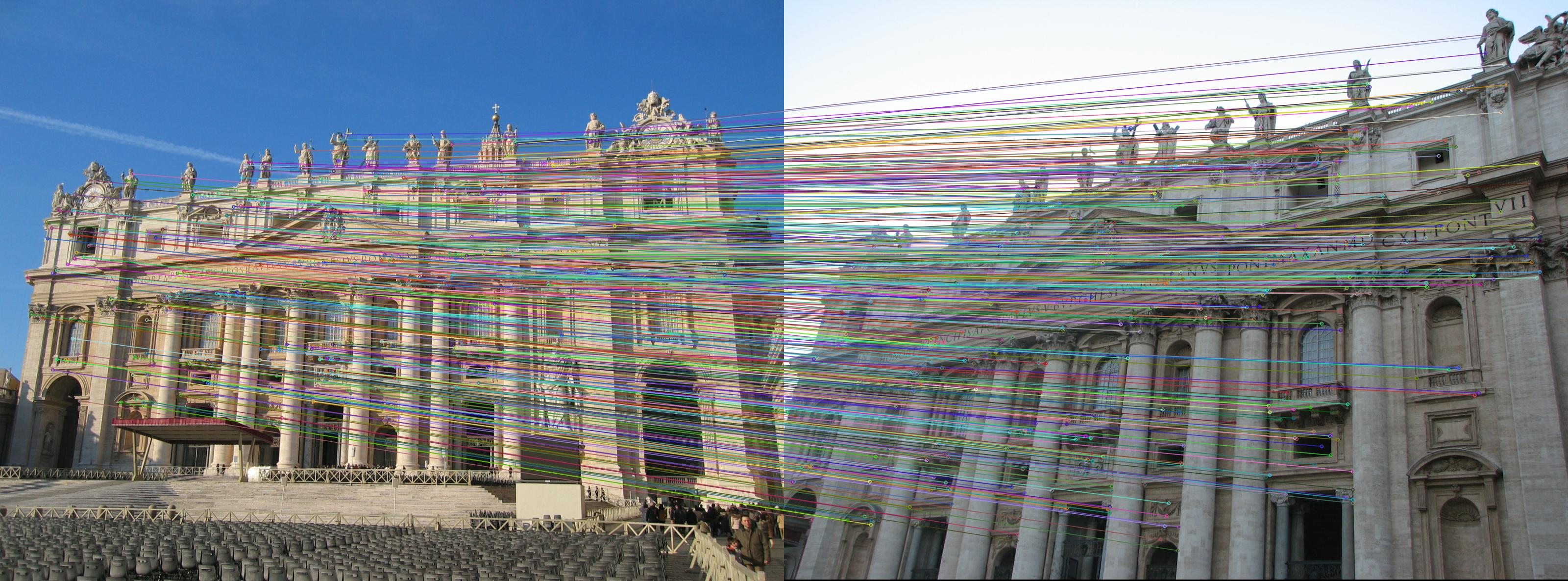}
    \caption{\textbf{Qualitative example of \ours~matches.} We picked a random pair in the MegaDepth-1500 testset and visualize confident \ours~matches. As can be seen in the figure, \ours~produces reliable keypoints and good matches, even in challenging situations. There are still outliers however, particularly for repeating structures such as similar statues, and identical pillars. Best viewed in high resolution.}
    \label{fig:qualitative-example-1}
\end{figure*}

\subsection{Inference}
For all SotA experiments we run at a resolution of $784\times 784$ and sample top-$k$ keypoints.

\subsection{Integrating \ours~With Matchers}
Recent works~\cite{sun2021loftr, edstedt2023dkm} often claim that detection localization errors hinder performance. To investigate this claim, we integrate \ours~into the recent SotA matcher RoMa~\cite{edstedt2023roma}.
We integrate RoMa into our detection framework by sampling the dense warp of RoMa at the location of \ours~keypoints, and match each keypoint by warping it to the other image and finding its nearest-neighbour DeDoDe keypoint there.

As expected, quantizing the RoMa warp to DeDoDe keypoints decreases performance compared to the results of RoMa, but increases the results compared to our detector+descriptor approach. If we view RoMa as an oracle, we can view the detector gap as about 7 points AUC@$5^{\circ}$. This indicates that there are still significant gains that can be made by improving the detector further.

\subsection{Qualitative Examples}
\parsection{\ours~Keypoints:}
Here we qualitatively demonstrate the types of keypoints and matches learned by \ours. In Figure~\ref{fig:qualitative-example-1} we show \ours~matches on a non-cherrypicked, \ie randomly chosen, pair from the MegaDepth-1500 testset. We find that \ours~produces repeatable keypoints that match well.

\subsection{SotA Comparison}
\parsection{MegaDepth-1500 Relative Pose:}
MegaDepth-1500 is a relative pose benchmark proposed in LoFTR~\cite{sun2021loftr}, and consists of 1500 pairs of images in two scenes of the MegaDepth dataset, which are non-overlapping with our training set. We compare \ours~against previous detector/descriptor models, as well as fully fledged matchers.
We present results in Table~\ref{tab:megadepth-loftr}. Our results clearly show the soundness of our approach, with gains of +17.8 AUC$@5^{\circ}$ compared to DISK, +12.9 AUC$@5^{\circ}$ compared to SiLK, and +10.9 AUC$@5^{\circ}$ compared to ALIKED.
 \begin{table}
 \small
     \centering
     \caption{\textbf{SotA comparison on the Megadepth-1500 benchmark}. The top portion contains dense and detector free matching methods, the middle part keypoint matching methods, while the bottom portion contains traditional MNN matching from keypoints and descriptors. Measured in AUC (higher is better).}
     \begin{tabular}{l lll}
     \toprule
      Method $\downarrow$\quad\quad\quad AUC $\rightarrow$& $@5^{\circ}$&$@10^{\circ}$&$@20^{\circ}$\\
      \midrule
          LoFTR~\cite{sun2021loftr}~\tiny{CVPR'21}& 52.8 & 69.2 & 81.2 \\
          SE2-LoFTR-4~\cite{bokman2022case}~\tiny{CVPRW'22} &  52.6 & 69.2 & 81.4 \\
          QuadTree~\cite{tang2022quadtree}~\tiny{ICLR'22} & 54.6	& 70.5	&82.2\\
ASpanFormer~\cite{chen2022aspanformer}~\tiny{ECCV'22} & 55.3  & 71.5 & 83.1\\
         ASTR~\cite{yuandchang2023astr}~\tiny{CVPR'23}  & 58.4 & 73.1 & 83.8\\
          DKM~\cite{edstedt2023dkm}~\tiny{CVPR'23}  &  60.4 & 74.9 & 85.1 \\
          PMatch~\cite{zhu2023pmatch}~\tiny{CVPR'23} & 61.4 & 75.7 & 85.7 \\
          RoMa~\cite{edstedt2023roma}~\tiny{Arxiv'23} &  \textbf{62.6} & \textbf{76.7} & \textbf{86.3}\\
          \textbf{DeDoDe} + RoMa &  55.1 & 71.6 & 83.5\\

          \midrule
          SuperGlue~\cite{sarlin2020superglue}~\tiny{CVPR'19} &  49.7 & \textbf{67.1} & 80.0 \\
          SGMNet~\cite{chen2021learning}~\tiny{CVPR'21} & 43.2 & 61.6 & 75.6\\
        LightGlue~\cite{lindenberger2023lightglue}~\tiny{ICCV'23} &  \textbf{49.9} & 67.0 & \textbf{80.1} \\
\midrule
SuperPoint~\cite{detone2018superpoint}~\tiny{CVPRW'18} & 31.7 & 46.8 & 60.1 \\
DISK~\cite{tyszkiewicz2020disk}~\tiny{NeurIps'20} & 35.0 & 51.4 & 64.9 \\
ALIKED~\cite{Zhao2023ALIKED}~\tiny{TIM'23} & 41.9 & 58.4 & 71.7 \\
SiLK~\cite{gleize2023silk}~\tiny{ICCV'23} & 39.9 & 55.1 & 66.9 \\
\textbf{DeDoDe-B} & 49.4 & 65.5 & 77.7\\
\textbf{DeDoDe-G} & \textbf{52.8} & \textbf{69.7} & \textbf{82.0}\\

     \bottomrule
     \end{tabular}

     \label{tab:megadepth-loftr}
 \end{table}
 
\parsection{MegaDepth Detector Repeatability:}
\begin{table}
\small
    \centering
    \caption{\textbf{SotA comparison on MegaDepth-Repeatability-10k.} *:Non-max supression removed. Note that \ours~and SiLK do not use non-max supression.}
    \begin{tabular}{l lll}
    \toprule
          Method & \multicolumn{3}{c}{Repeatability @ 10k} \\  
          &$0.1\%$&$0.2\%$&$0.5\%$\\
          \midrule
DISK~\cite{tyszkiewicz2020disk}~\tiny{Neurips'20} & 18.4 & 56.6 & \textbf{93.0} \\
DISK*~\cite{tyszkiewicz2020disk}~\tiny{Neurips'20} & 32.6 & 60.0 & 83.9 \\
ALIKED~\cite{Zhao2023ALIKED}~\tiny{IEEE-TIM'23} & 19.4 & 63.3 & 90.8 \\
ALIKED*~\cite{Zhao2023ALIKED}~\tiny{IEEE-TIM'23} & 26.4 & 63.2 & 85.6 \\
SiLK~\cite{gleize2023silk}~\tiny{ICCV'23} & 21.2 & 46.5 & 67.1 \\
\textbf{\ours} & \textbf{40.1} & \textbf{63.7} & 80.3 \\
\bottomrule
    \end{tabular}
    \label{tab:detect-repeatability-10000}
\end{table}
\begin{table}
\small
    \centering
        \caption{\textbf{SotA comparison on MegaDepth-Repeatability-2k.} *: Non-max supression removed. Note that \ours~and SiLK do not use non-max supression.}
    \begin{tabular}{l lll}
    \toprule
          Method & \multicolumn{3}{c}{Repeatability @ 2k} \\
          &@$0.1\%$&@$0.2\%$&@$0.5\%$\\
          \midrule
DISK~\cite{tyszkiewicz2020disk}~\tiny{Neurips'20} & 17.0 & 41.1 & \textbf{65.7} \\
DISK*~\cite{tyszkiewicz2020disk}~\tiny{Neurips'20} & 21.8 & 40.1 & 57.5\\
SiLK~\cite{gleize2023silk}~\tiny{ICCV'23} & 8.2 & 21.8 & 35.8 \\
ALIKED~\cite{Zhao2023ALIKED}~\tiny{IEEE-TIM'23} & 19.3 & 45.4 & \textbf{67.9} \\
ALIKED*~\cite{Zhao2023ALIKED}~\tiny{IEEE-TIM'23} & 21.2 & \textbf{46.1} & 66.5 \\
\textbf{\ours} & \textbf{24.3} & 41.8 & 58.3 \\
\bottomrule
    \end{tabular}
    \label{tab:detect-repeatability-2000}
\end{table}
Previous benchmarks for keypoint detectors often jointly measure the repeatability of keypoints and the matchability of the descriptors into single metrics. Here we aim to disentangle these two measurements, and \emph{only} measure keypoint repeatability. To do this, we use the ground truth warp, and measure the percentage of inliers, \ie, pairs of keypoints close enough, under a set of thresholds.
We investigate two different settings, the many keypoints setting (maximum 10000 keypoints), and the few keypoint setting (maximum 2000 keypoints).
Results are presented in Table~\ref{tab:detect-repeatability-10000} and Table~\ref{tab:detect-repeatability-2000}. The results reveal interesting properties of detectors.
In particular, we find that \ours~and SiLK show large gains from higher numbers of keypoints for tight thresholds, while DISK shows large gains in loose thresholds.
This indicates that DISK detections consists of a small number of certain detections, with a large number of uncertain detections, whereas \ours~and SiLK detections contain a larger spectrum.
This might be due to the training of DISK, which is regularized to have single detections in patches by local softmax~\cite{tyszkiewicz2020disk}, while \ours~and SiLK do not.
It can also be observed that ALIKED shows very minor gains from increasing the number of keypoints.
This indicates that ALIKED may be better suited for usecases where very few but reliable keypoints are required, while \ours~scales better for larger numbers. Finally, note that while \ours, ALIKED, and DISK are trained on the MegaDepth dataset, SiLK is trained only using synthetic homographies.

\parsection{Image Matching Challenge 2022:}
The Image Matching Challenge 2022~\cite{image-matching-challenge-2022} is composed of challenging uncalibrated relative pose estimation pairs. Different from MegaDepth-1500, the test set is hidden and does not derive from MegaDepth, and may therefore better indicate generalization performance, especially for models such as \ours~that train on MegaDepth.
We follow the setup in SiLK~\cite{gleize2023silk} and use \num{30000} keypoints, and MAGSAC++~\cite{barath2020magsac++} with a threshold of $0.25$ pixels. We use a fixed image size of $784\times 784$. We follow the approach in RoMa~\cite{edstedt2023roma} and report results on the hidden test set. We present results in Table~\ref{tab:imc2022}. Here too, \ours~achieves significant improvements to the state-of-the-art, with results competitive with the graph neural network (GNN) 
 matcher SuperGlue~\cite{sarlin2020superglue}, and surpassing the recent SiLK~\cite{gleize2023silk} with +7.4 points mAA.
\begin{table}
\small
    \centering
    \caption{\textbf{SotA comparison on the IMC2022 benchmark.} Relative pose estimation results on the IMC2022~\cite{image-matching-challenge-2022} hidden test set, measured in mAA (higher is better). The top portion contains dense and detector free matching methods, the middle part keypoint matching methods, while the bottom portion contains traditional MNN matching from keypoints and descriptors.}    \label{tab:imc2022}
    \begin{tabular}{ll}
    \toprule
     Method $\downarrow$\quad\quad\quad mAA $\rightarrow$&$@10$\\
     \midrule
LoFTR~\cite{sun2021loftr}~\tiny{CVPR'21} & 78.3 \\
MatchFormer~\cite{wang2022matchformer}~\tiny{ACCV'22} &78.3 \\
QuadTree~\cite{tang2022quadtree}~\tiny{ICLR'22} &81.7 \\
ASpanFormer~\cite{chen2022aspanformer}~\tiny{ECCV'22} &83.8    \\
RoMa~\cite{edstedt2023roma}~\tiny{Arxiv'23} & \textbf{88.0}\\
\midrule
SP~\cite{detone2018superpoint}+SuperGlue~\cite{sarlin2020superglue} &\textbf{72.4} \\
\midrule
DISK~\cite{tyszkiewicz2020disk}~\tiny{Neurips'20} & 64.8 \\
ALIKED~\cite{Zhao2023ALIKED}~\tiny{IEEE-TIM'23} & 64.9 \\
SiLK~\cite{gleize2023silk}~\tiny{ICCV'23} & 68.5 \\
\textbf{DeDoDe-B} & \textbf{72.9} \\
\textbf{DeDoDe-G} & \textbf{75.8} \\

\bottomrule
    \end{tabular}
\end{table}

\subsection{Ablation Study}

\parsection{Design Decisions:}
\begin{table}
\small
    \centering
    \caption{\textbf{Ablation study on MegaDepth-Repeatability-10k.} We investigate the performance effect of two of our contributions on the detector repeatability using a fixed resolution of $512\times 512$.}
    \begin{tabular}{l lll}
    \toprule
          Method & \multicolumn{3}{c}{Repeatability @ 10k} \\  
          &$0.1\%$&$0.2\%$&$0.5\%$\\
          \midrule
\label{case:I}{\textcolor{red}{I}}: Prior Distribution as Target &34.8 & 61.3 & 78.0\\
\label{case:II}{\textcolor{red}{II}}: No Coverage Loss & 29.7& 59.4& 78.0\\
\textbf{\ours} (512x512) & \textbf{37.1} & \textbf{68.1} & \textbf{84.8} \\
\bottomrule
    \end{tabular}
    \label{tab:ablation-repeatability-10000}
\end{table}
Here we investigate the impact of our design decisions. In Setting \hyperref[case:I]{I}, we supervise directly using the detection prior in Section~\ref{sec:prior-detection-dist}. This decreases performance, in particular for coarse thresholds. In Setting \hyperref[case:II]{II} we do not include the coverage term introduced in Section~\ref{sec:detector-training-objective}. This decreases performance across all thresholds. We attribute this to the network not learning to attenuate certainty in unmatchable regions, \eg, sky.

\parsection{Exchanging Components with Previous Methods:}
In this section we compare the effect of the detector and descriptor of \ours~by exchanging components of SIFT, DISK, and \ours. We present results in Table~\ref{tab:ablation}. Interestingly, using SIFT detections with \ours~descriptions outperforms the DISK feature matcher. Similarly, replacing the DISK descriptor with \ours~also improves performance. Note however that while results are improved compared to the original methods with \ours~descriptor, it is a significant decrease compared to when using \ours~detections, with an absolute decrease of 8 points AUC.

Replacing the detector in DISK with \ours~slightly decreases the performance. We attribute this to the training of DISK, which learns keypoints that is likely to be mutual nearest neighbours of the descriptor. 
SIFT descriptors, in contrast to DISK, require scale and rotation estimation from the keypoint. To circumvent this issue we sample SIFT keypoints within a distance of $1$ pixel from \ours~keypoints. Note that this is a inferior version to our original detector as it results in a subset of the original SIFT keypoints. It is therefore not surprising that results are slightly degraded.
 \begin{table}
 \small
     \centering
     \caption{\textbf{Ablation study on the Megadepth-1500 benchmark}. Here we ablate the gains from incorporating the DeDoDe detector and descriptor to matching. As baselines we compare to using SIFT, or DISK~\cite{tyszkiewicz2020disk}.
     $^*$: SIFT requires scale and orientation estimation from the keypoints, see text for details.}
     \begin{tabular}{l lll}
     \toprule
      Detector/Descriptor $\downarrow$\quad AUC $\rightarrow$& $@5^{\circ}$&$@10^{\circ}$&$@20^{\circ}$\\
      \midrule
SIFT/SIFT~\cite{lowe2004distinctive} & 36.5 & 50.2 & 62.7 \\
DISK/DISK~\cite{tyszkiewicz2020disk}~\tiny{NeurIps'20} & 35.0 & 51.4 & 64.9 \\
\textbf{DeDoDe/DeDoDe-B} & \textbf{49.4} & \textbf{65.5} & \textbf{77.7}\\
\midrule
SIFT/DeDoDe-B & 41.1 & 56.8 & 69.8\\
DISK/DeDoDe-B & 41.5 & 57.9 & 71.0\\
\midrule
DeDoDe/SIFT$^*$ & 34.1 & 49.1 & 62.0\\
DeDoDe/DISK & 33.1 & 48.9 & 61.7\\
     \bottomrule
     \end{tabular}

     \label{tab:ablation}
 \end{table}
\section{Conclusion}
We have presented \ours, a descriptor-agnostic modular approach to geometric keypoint detection. We did this by formulating keypoint detection as detecting points that have successfully created 3D tracks in large-scale SfM. We introduced a two-view objective, in order to reduce the sparseness of the detections, and furthermore proposed a suitable descriptor objective. Our experiments show that \ours~signficantly improves on the state-of-the-art. \ours~enables a new research direction towards a modular detection and matching framework.

\parsection{Limitations:}
\begin{enumerate}
    \item \ours~is dependent on the base detector. This means that potential stable keypoints can be missed. We remedied this by introducing a top-$k$ semi-supervised objective during training.
    \item \ours~is not trained with augmentations, contrary to common practice~\cite{lindenberger2023lightglue,gleize2023silk}. It is possible that appropriate augmentation would increase the performance/generalization of our method.
    \item \ours~tends to produce a large number of potential keypoints. In the few-keypoint setting this means that the repeatability degrades, as many keypoints are missed. Adjusting the hyperparameters of the keypoint target distribution may be able to address these issues. 
    \item The \ours~detector produces only keypoint locations, and does not estimate orientation/scale. This could potentially be remedied by using a separate network for estimating the local frame~\cite{mishkin2018repeatability,lee2021self,yan2022learning}, although this would come at an additional computational cost.
\end{enumerate}

\paragraph{Acknowledgements:} 
We thank the reviewers for the constructive feedback that helped improve the paper.
We thank Emanuel Sanchez Aimar and Ioannis Athanasiadis for the early discussions.
This work was supported by the Wallenberg Artificial
Intelligence, Autonomous Systems and Software Program
(WASP), funded by Knut and Alice Wallenberg Foundation; and by the strategic research environment ELLIIT funded by the Swedish government. The computational resources were provided by the
National Academic Infrastructure for Supercomputing in
Sweden (NAISS) at C3SE partially funded by the Swedish Research
Council through grant agreement no.~2022-06725, and by
the Berzelius resource, provided by the Knut and Alice Wallenberg Foundation at the National Supercomputer Centre.

\newpage
{
    \small
    \bibliographystyle{ieeenat_fullname}
    \bibliography{main}
}
\clearpage
\maketitlesupplementary
\appendix

In this supplementary material we present implementation details, qualitative examples, and theory that could not fit in the main text.
\section{Definitions of Evaluation Metrics}

\parsection{AUC:} We evaluate performance using the Area Under the Curve (AUC). We define the AUC$@t$ as the integral of the precision up to the threshold $t$. In practice, this is approximated using the trapezoidal rule, as in previous work~\cite{sun2021loftr,edstedt2023dkm}.

\parsection{Precision:} The relative pose estimation precision is defined as 
\begin{equation}
    {\rm Precision}@t = \max (|\varepsilon_{\hat{t}}|,|\varepsilon_{\hat{R}}|) < t
\end{equation}
where 
\begin{equation}
 \varepsilon_{\hat{t}} = \arccos \frac{\langle t, \hat{t}\rangle}{\lVert t \rVert \cdot \lVert \hat{t} \rVert}
\end{equation}
and

\begin{equation}
 \varepsilon_{\hat{R}} = \arccos\big( ({\rm tr} (\hat{R}^{\top} R) - 1)/2 \big).
\end{equation}

\parsection{Repeatability:}
First, let
\begin{equation}
    \textbf{x}^{\A}_{\rm kp \cap W} = \{x^{\A}_{i} | x^{\A}_{i}\in {\rm dom}(W) \cap \textbf{x}^{\A}_{\rm kp}\},
\end{equation}
where W is the ground truth warp, and its domain {\rm dom}(W) are the pixels where it is well defined.
In the case of MegaDepth, we will only consider keypoints in the MVS regions. We define repeatability as 
\begin{equation}
    \frac{1}{|\textbf{x}^{\A}_{\rm kp \cap W}|}\sum_{x^{\A}_{i}\in\textbf{x}^{\A}_{\rm kp \cap W}} 
    \bigg(\min_{x^{\B}_{i}\in\textbf{x}^{\B}_{\rm kp}}
    \lVert x^{\B}_{i}-W(x^{\A}_{i})\rVert_2 \bigg) < t.
\end{equation}
In words, this measures the the percentage of keypoints detections in the MVS covisible region with a corresponding keypoint within a certain threshold in the other image.

\parsection{mAA:}
For IMC-2022 the performance is measured in mAA, which closely resembles AUC. We further detail the metric below.
Given a an estimated fundamental matrix, the error is computed in terms of rotation ($\varepsilon_R$ in degrees) and translation ($\varepsilon_T$ in meters). Given one threshold over each, the pose is classified as accurate if it meets both thresholds. In code:\\
{\tt thresholds\_r = np.linspace(1, 10, 10) \# In degrees.}\\ 
{\tt thresholds\_t = np.geomspace(0.2, 5, 10)  \# In meters.}\\
The percentage of image pairs that meet every pair of thresholds is computed, the results are averaged over all thresholds, which rewards more accurate poses. As the dataset contains multiple scenes, which may have a different number of pairs, the metric is computed separately for each scene and then averaged. We refer to this metric as mAA$@10$.

\section{\ours~Further Implementation Details}
In the following paragraphs, we describe implementation details that did not fit in the main text.

\parsection{Detector Objective Further Details:}
We set the prior Uniform constant $C$ such that the peak of the Gaussian centered around the deltas is a factor $\exp(50)$ larger compared to the uniform. In practice, we do these computations in log-space. This ensures that all consistent keypoints are detected, while still allowing the network to detect additional keypoints in cases where the target distribution is overly sparse. We found that the performance of the network was robust to the precise value of $C$.

\parsection{Further Detector Architecture Details:}
For \ours~detector dimensionality of the encoded context is $[32, 128, 256]$ for strides $[2,4,8]$.

\parsection{Descriptor Objective Further Details:}
In addition to enforcing the keypoints to be mutual nearest neighbours, we enforce that the distance be smaller than $0.5\%$ of the image in both directions. We did not find that this had a major impact on performance, possibly due to very few mutual nearest neighbours not fulfilling the criteria.

\parsection{Descriptor Architecture Further Details:}
For \ours~descriptor-B dimensionality of the encoded context is $[32, 128, 256]$ for strides $[2,4,8]$ respectively, for \ours~descriptor-G we additionally encode context of dimension $512$ for the DINOv2~\cite{oquab2023dinov2} features at stride $14$.

\parsection{Training Further Details:}
We use a base learning rate of $2\cdot 10^{-5}$ for the encoders and $1\cdot 10^{-4}$ for the decoders. We use a cosine decay learning rate scheduler.

\section{Runtime:}
Here we measure the runtime of \ours, and compare it to DISK~\cite{tyszkiewicz2020disk}. We run both methods at the inference resolution used in the SotA experiments, which is $1024\times 1024$ for DISK and $784\times 784$ for \ours. We present results in Table~\ref{tab:runtime}. We find that \ours~has a similar runtime to DISK, and is about $\sim 2.5\times$ faster for keypoint-only extraction. 
\begin{table}
    \centering
    \begin{tabular}{ll}
    \toprule
    & Time (ms) \\
         \midrule
        \ours~Det&  62.5\\
        \ours~Det+Desc-B&  103.4\\
        \ours~Det+Desc-G&  160.0\\
        DISK Det+Desc & 170.0 \\
         \bottomrule
    \end{tabular}
    \caption{\textbf{Inference time comparison.} We compare the inference time of DeDode to DISK.}
    \label{tab:runtime}
\end{table}
\section{Inference Settings for Detector+Descriptors}
\parsection{\ours:} We use a resolution of $784\times 784$ We use a dual-softmax threshold of 0.01.

\parsection{DISK:} We follow the suggested inference approach, and use images of size $1024\times 1024$, with a NMS radius of 5. We use a dual-softmax matcher with confidence threshold $0.01$. We additionally investigated the ratio-test but found it to give inferior results on IMC22.

\begin{figure*}
    \centering
    \includegraphics[width=.48\linewidth]{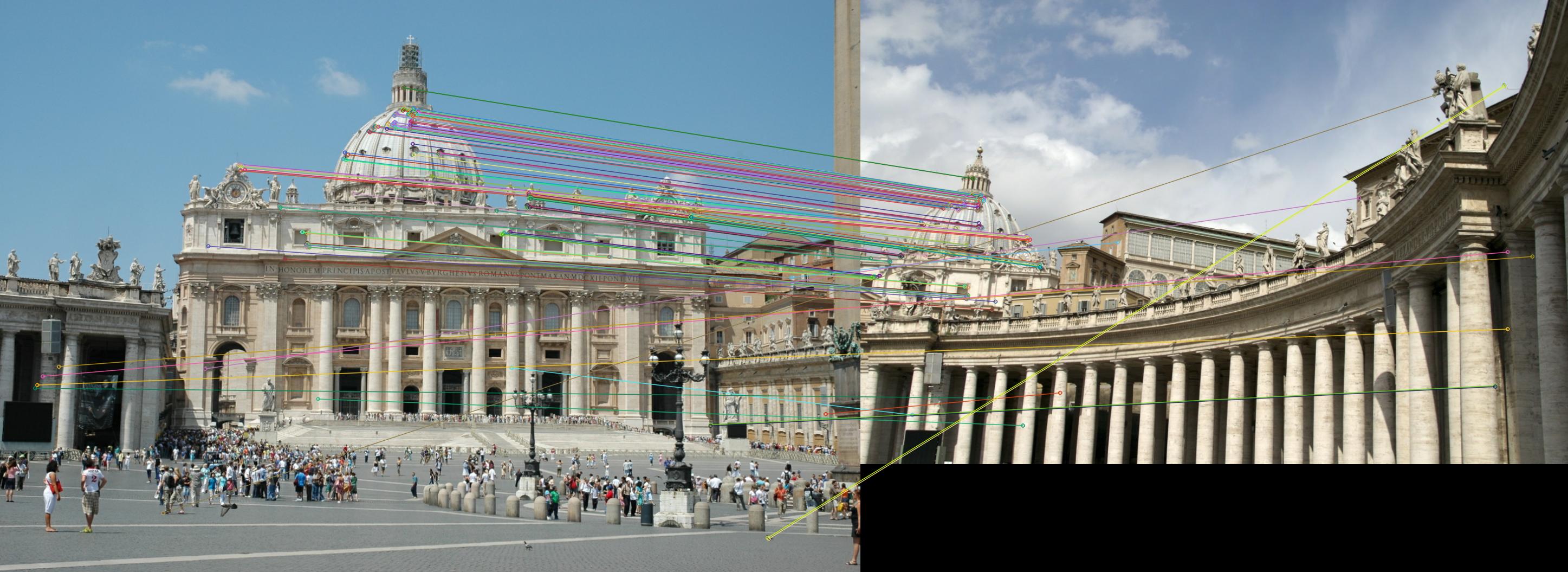}
    \includegraphics[width=.48\linewidth]{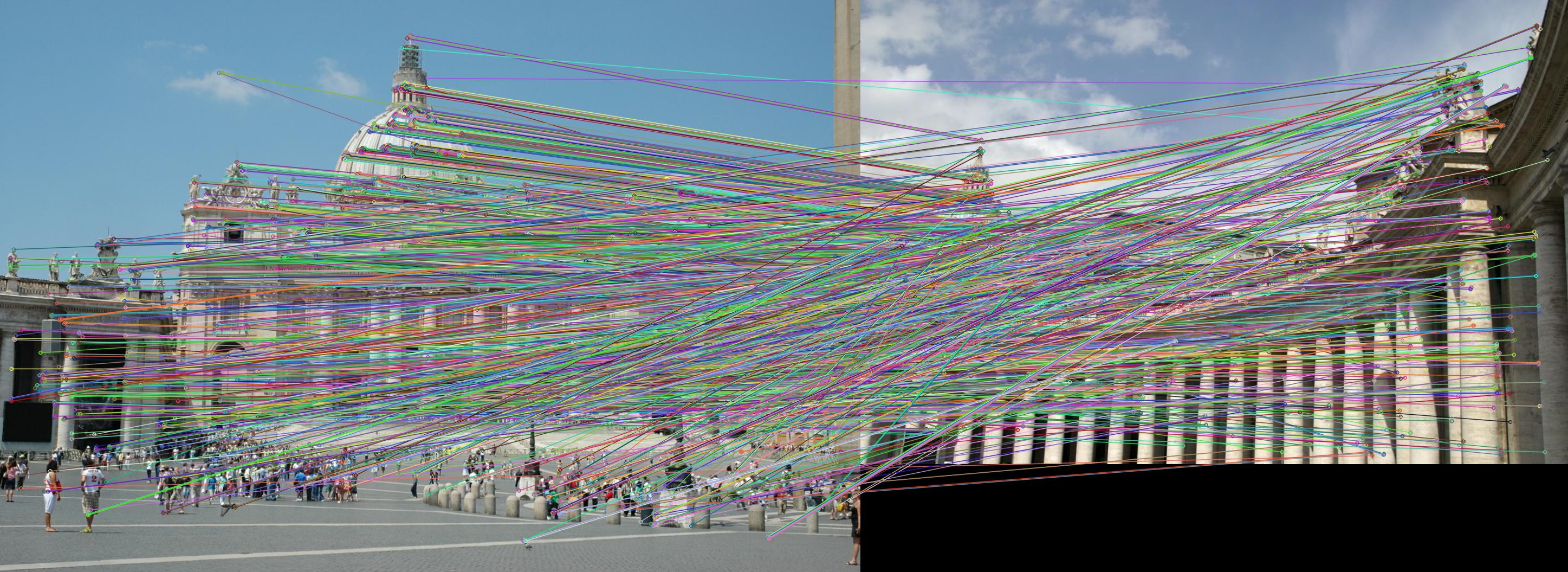}
    
    \caption{\textbf{Issues with SiLK confidence.} At tight confidence thresholds SiLK tends to produce too few but accurate correspondences. However, at looser thresholds a large number of outlier correspondences arise.}
    \label{fig:silk-conf}
\end{figure*}

\parsection{SiLK:} We use the \href{https://www.kaggle.com/code/piezzo/imc2022-submission-silk/notebook}{ official settings of SiLK for IMC-2022}, where the longer side of the image is resized to $720$, a dual-softmax matcher with confidence threshold of $0.01$, and use \num{30000} keypoints. Using this threshold, SiLK produces many outliers, as can be seen in Figure~\ref{fig:silk-conf}. For a fair comparison on MegaDepth-1500 we therefore instead use a threshold of $0.05$, which we found beneficial for SiLK.

\parsection{ALIKED:} We find ALIKED to perform the best using a resolution of $800\times 800$, we sample \num{10000} keypoints per image, and use a dual-softmax matcher with confidence threshold of $0.01$.
\begin{figure}
    \centering
    \includegraphics[width=.8\linewidth]{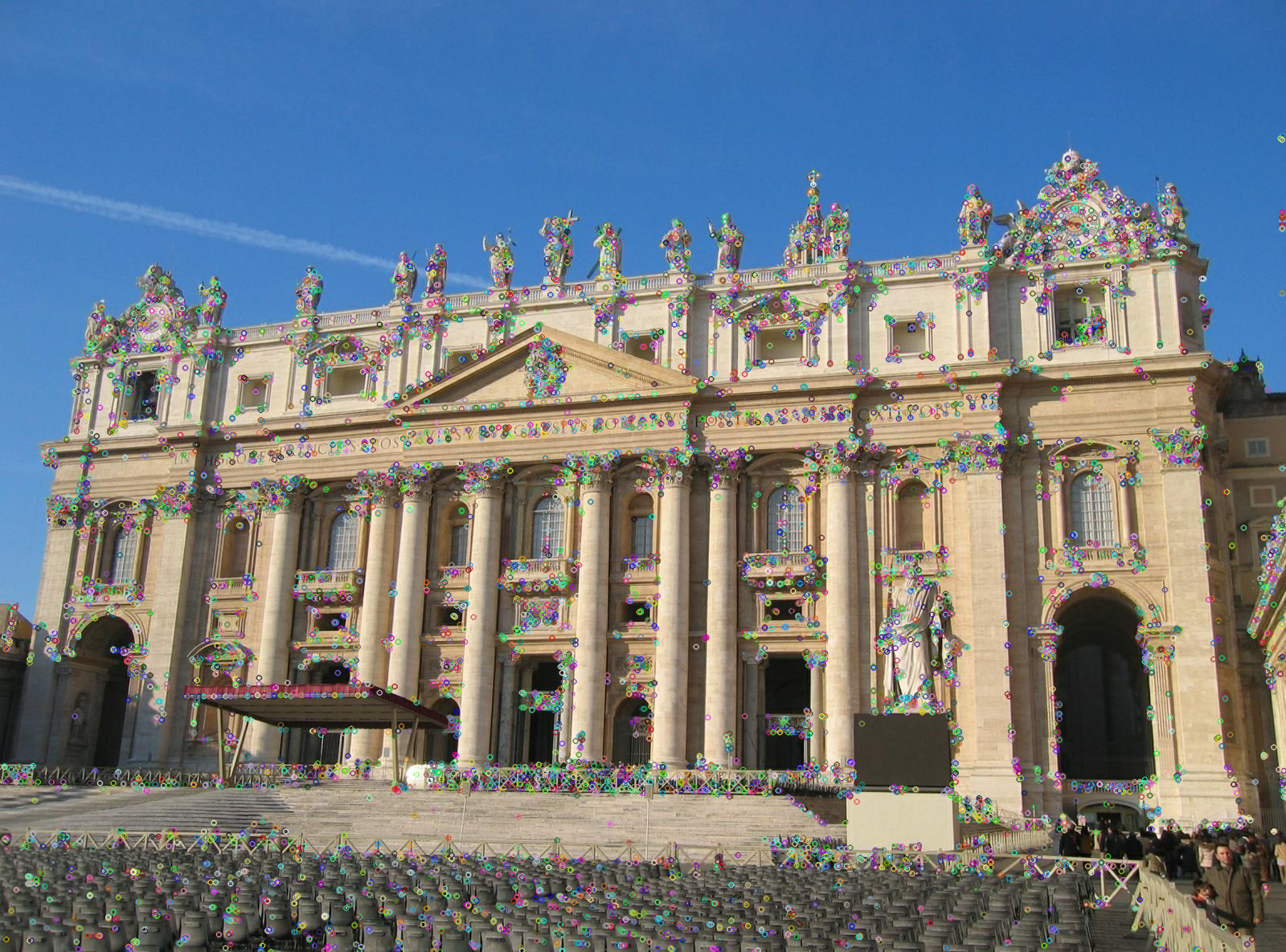}
    \caption{\textbf{Top-10k Keypoints of \ours.} In contrast to previous methods, our method is able to produce a larger amount of repeatable keypoints, leading to more accurate estimation results. Best viewed in high-resolution.}
    \label{fig:logit-vis}
\end{figure}

\section{Robustness to Large Rotations}
We revisit the pair in Figure~\ref{fig:qualitative-example-1} in a more difficult setting by rotating the first image by $180^{\circ}$. 
We find that despite not being trained on large rotations, \ours~produces good matches. We leave investigations of how to improve performance on non-upright images to future work.

\begin{figure}
    \centering
    \includegraphics[width=\linewidth]{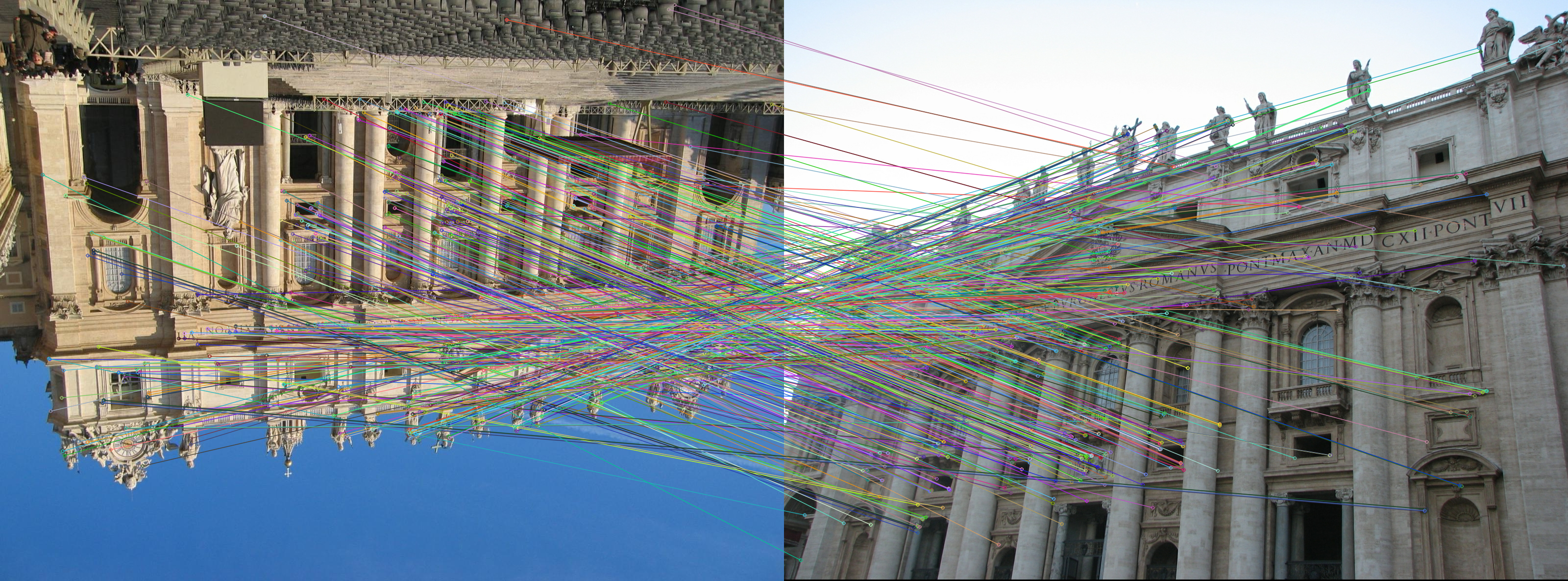}
    \caption{\textbf{Robustness of \ours~to large rotation.} Despite not being trained on large rotations, and not having an equivariant network architecture, \ours~is able to match under even extreme changes in rotation. Rotations do however produce a larger number of outlier correspondences as can be seen in the figure.}
    \label{fig:qualitative-rot}
\end{figure}

\begin{figure*}
    \centering
    \includegraphics[width=.49\linewidth]{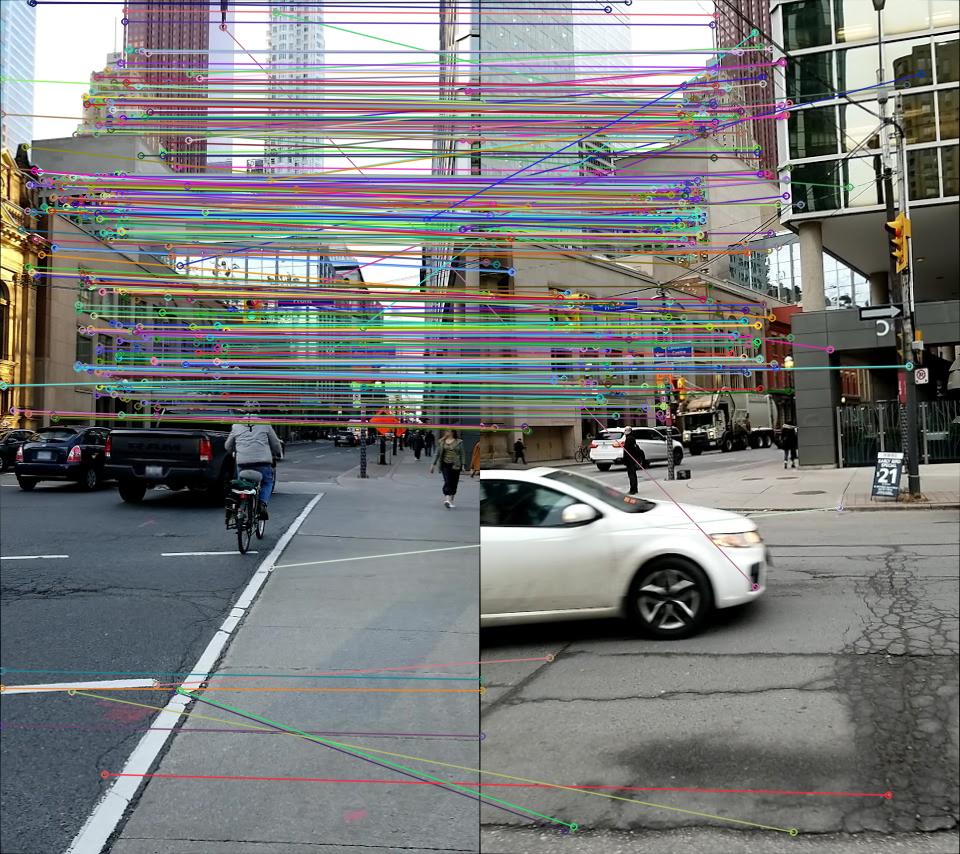}
    \includegraphics[width=.49\linewidth]{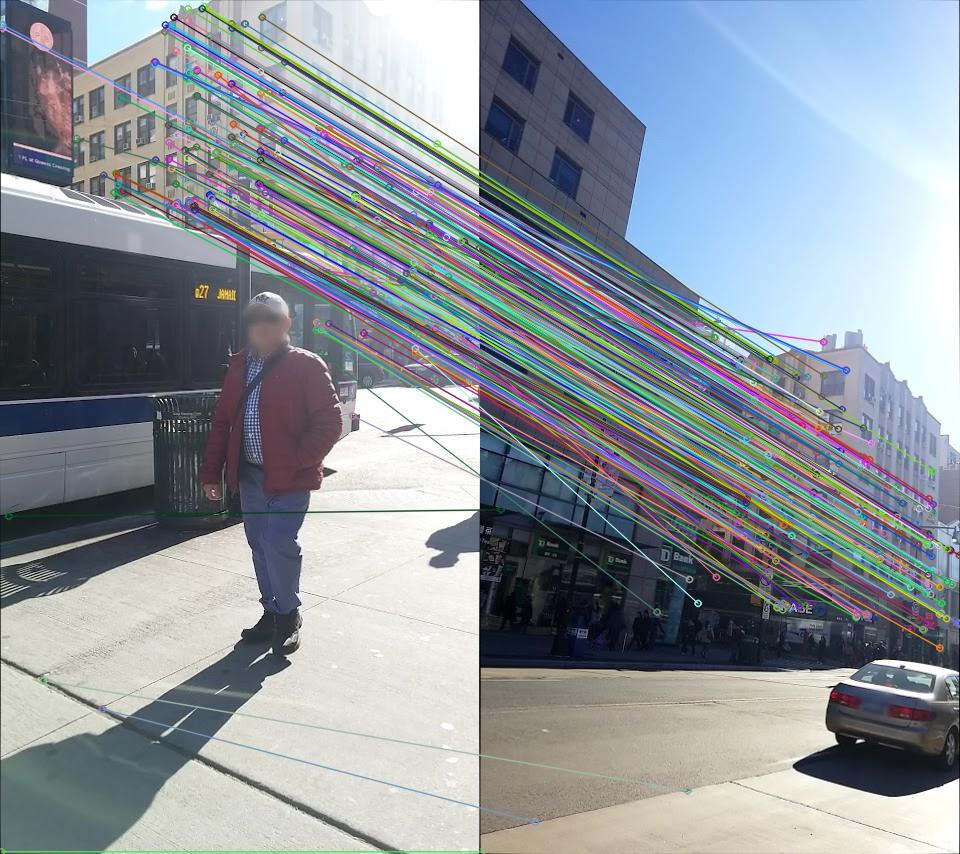}
    \caption{\textbf{\ours~matches on IMC-2022.} We show matches of two publicly available image pairs from the IMC2022 benchmark.}
    \label{fig:dedode-imc22}
\end{figure*}

\begin{figure*}
    \centering
    \includegraphics[width=\linewidth]{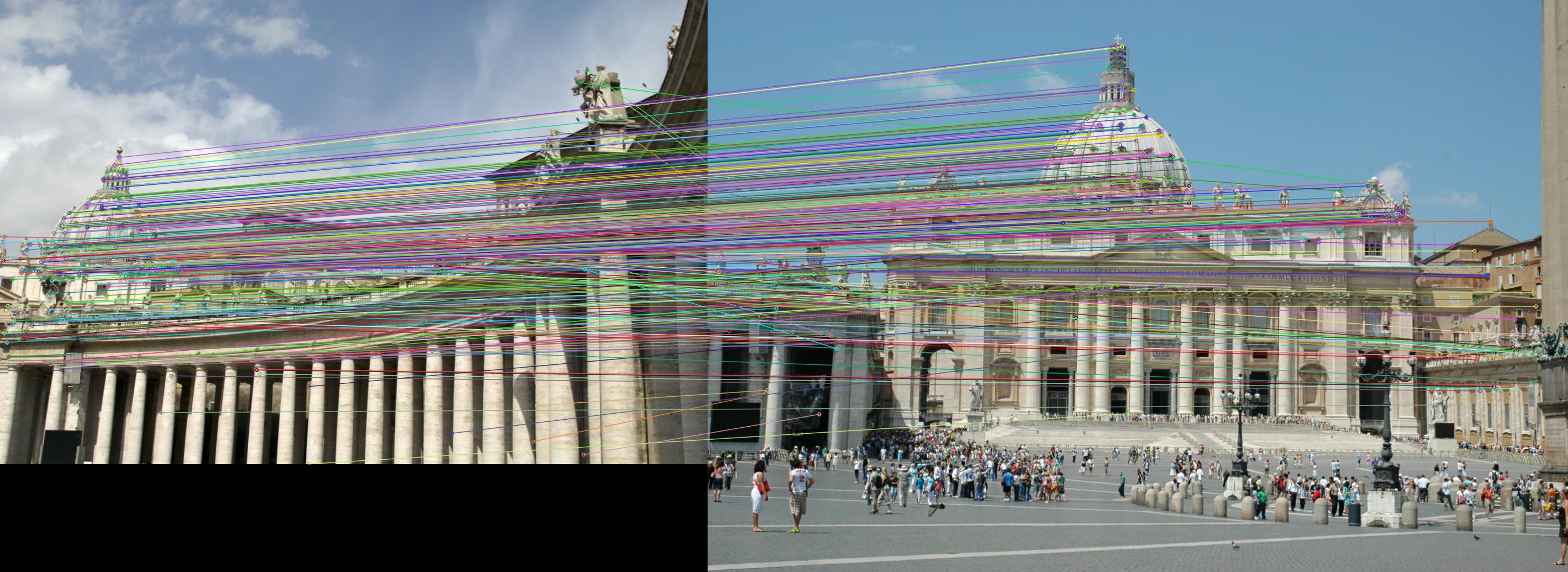}
    \caption{\textbf{\ours~matches on the MegaDepth-1500 benchmark.} }
    \label{fig:dedode-mega0015}
\end{figure*}

\section{Why is top-$k$ Needed?}
\label{sec:why-topk}
In Section~\ref{sec:detector-training-objective} we argued that the top-k thresholding is needed to stabilize training. Here we provide some theoretical reasons why the non-thresholded objective has issues.

To see this, note that the cross-entropy is 0 if and only if $p_{f_{\theta}} = \delta_{x_i}(x)$, for any $x_i$. Hence this objective has multiple degenerate solutions. We found that when training, the network collapses into predicting a single peak.

One could instead consider a Kullback-Leibler (KL) divergence between the distributions. This divergence is however also minimized by any $p_{f_{\theta}} = \delta_{x_i}(x)$.

Stability issues in self-supervised objectives are well-known and many solutions have been proposed, see \eg,~\citet{grill2020bootstrap, oquab2023dinov2}.

Our proposed top-$k$ objective is a simple way of avoiding these instabilities. The binarization encourages the network to predict a flat distribution over certain keypoints, ensuring that it does not collapse. The downside of the objective is, as discussed in Section~\ref{sec:detector-training-objective} that $k$ is an additional hyperparameter that is not obvious how to set optimally and will depend on the dataset.

\section{Additional Qualitative Examples}
In Figure~\ref{fig:dedode-imc22} we provide qualitative examples of \ours~on the publically released images of the IMC-2022 benchmark. In Figure~\ref{fig:dedode-mega0015} we provide an additional example of \ours~on the MegaDepth-1500 benchmark.

\end{document}